\documentclass[10pt,twocolumn,letterpaper]{article}

\usepackage{cvpr}
\usepackage{times}
\usepackage{epsfig}
\usepackage{graphicx}
\usepackage{amsmath}
\usepackage{amssymb}

\usepackage{blindtext}
\usepackage{xcolor}
\usepackage{booktabs}
\usepackage{makecell}
\usepackage{multirow}
\usepackage{amsmath}
\usepackage{subfigure}
\usepackage{fancyhdr}
\usepackage[final]{pdfpages}

\DeclareMathOperator*{\argmax}{arg\,max}

\usepackage[pagebackref=true,breaklinks=true,letterpaper=true,colorlinks,bookmarks=false]{hyperref}

\cvprfinalcopy 

\def\cvprPaperID{9054} 

\ifcvprfinal\fi
\begin{document}

\title{Straight to the Point: Fast-forwarding Videos via Reinforcement \\Learning Using Textual Data}


\author{Washington Ramos\textsuperscript{1} \hspace{.3em} Michel Silva\textsuperscript{1} \hspace{.3em} Edson Araujo\textsuperscript{1} \hspace{.3em} Leandro Soriano Marcolino\textsuperscript{2}\\ Erickson Nascimento\textsuperscript{1}\\
	\textsuperscript{1}Universidade Federal de Minas Gerais (UFMG), Brazil \hspace{.5em} \textsuperscript{2}Lancaster University, UK\\
	{\tt\footnotesize \textsuperscript{1}\{washington.ramos, michelms, edsonroteia, erickson\}@dcc.ufmg.br, \textsuperscript{2}l.marcolino@lancaster.ac.uk}
}

\maketitle
\thispagestyle{fancy}
\fancyhf{}
\chead{{To appear in Proceedings of the 2020 IEEE/CVF Conference on Computer Vision and Pattern Recognition (CVPR) \\ The final publication will be available soon.}}

\begin{abstract}

The rapid increase in the amount of published visual data and the limited time of users bring the demand for processing untrimmed videos to produce shorter versions that convey the same information. Despite the remarkable progress that has been made by summarization methods, most of them can only select a few frames or skims, which creates visual gaps and breaks the video context. In this paper, we present a novel methodology based on a reinforcement learning formulation to accelerate instructional videos. Our approach can adaptively select frames that are not relevant to convey the information without creating gaps in the final video. Our agent is textually and visually oriented to select which frames to remove to shrink the input video. Additionally, we propose a novel network, called Visually-guided Document Attention Network (VDAN), able to generate a highly discriminative embedding space to represent both textual and visual data. Our experiments show that our method achieves the best performance in terms of F1 Score and coverage at the video segment level.

\end{abstract}

\section{Introduction}
\label{sec:introduction}

From the dawn of the digital revolution until this very day, we are witnessing an exponential growth of data, in particular, textual and visual data such as images and videos. New technologies like social media and smartphones massively changed how we exchange and acquire information. For instance, there is a plethora of textual tutorials and instructional videos on the Internet teaching a variety of tasks, from how to cook burritos and tacos, all the way to how to solve partial differential equations (PDEs), and to device operations manuals.

Despite many textual tutorials and instructional videos sharing the increasing growth of available data as well as the content, they differ in a key aspect for users: {\it how long they would take to consume the content}. In general, information encoded by producers is more concise in textual data than when they use visual data. For instance, a recipe of tacos or a tutorial explaining how to solve a PDE is described in a few sentences. Instructional videos, for their turn, might have several minutes showing non-relevant information for the task such as a person opening a refrigerator, picking up the pencil, or erasing the blackboard. Such segments could be fast-forwarded without losing the crucial information encoded in the input video to understand the task. Thus, ideally, instructional videos should be concise, similar to a textual description, but still having visually-rich demonstrations of all main steps of the task.

\begin{figure}
	\centering
	\includegraphics[width=.95\columnwidth]{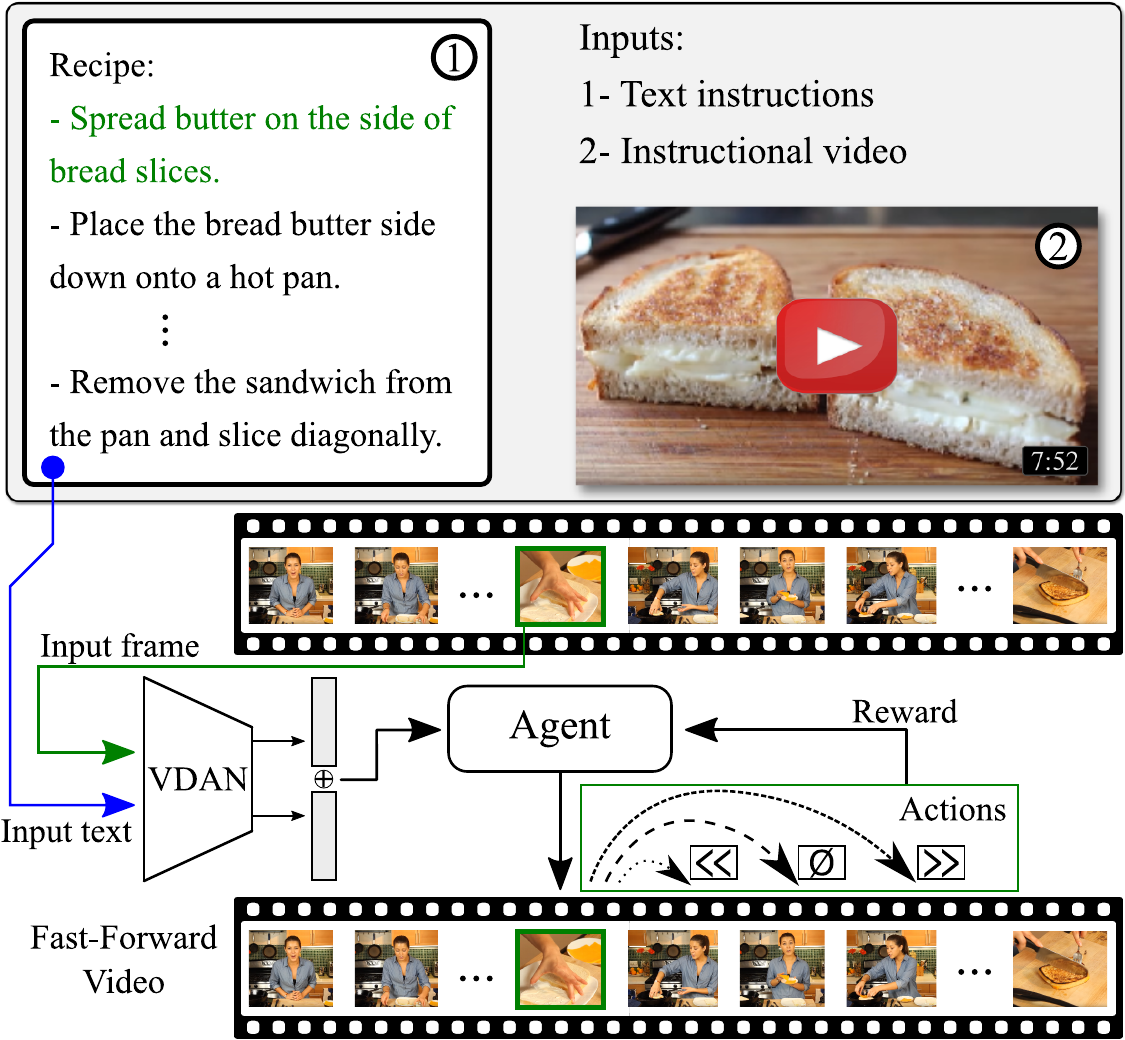}
	\caption{Schematic diagram of our fast forward method. After creating an embedding space for encoding the document and video frames, we train an agent that observes the encoded text and frame and chooses an action (\eg, increase, decrease or keep the speed-up rate) to emphasize highly semantic segments of the input video.}
	\label{fig:intro}
\end{figure}

In this paper, we address the problem of accelerating untrimmed videos by using text documents (see Figure~\ref{fig:intro}). For example, a recipe for cooking tacos could be used as a guide to select relevant frames from cooking tacos videos. Note that this problem is different from video segmentation~\cite{Zhou2018Towards} or summarization~\cite{Lee2012, Fei2017, Mahasseni2017, Zhang2016} since non-relevant frames are still necessary for a user to understand the flow and temporal coherence of a task, \ie, some segments should be accelerated, but not eliminated.

Our method follows the encoding-decoding framework to create fast-forward videos guided by a set of sentences (\ie, a document). We formulate our fast-forwarding task as a sequential decision-making process, where a reinforcement learning agent observes the encoded text and video frames and decides to increase, decrease, or keep the speed-up rate of the video. The embedding space for user documents and video frames is generated by a novel Visually-guided Document Attention Network (VDAN), which creates representative feature vectors for textual and visual modalities. In this embedding space, the vectors will be close when representing the same semantic concept and distant otherwise. Figure~\ref{fig:intro} shows a schematic representation of the main steps in our methodology.

Despite remarkable advances in summarization methods \cite{Lee2012, Fei2017, Mahasseni2017, Zhang2016}, most existing approaches do not take into account the temporal continuity, \ie, summarization techniques segment the input videos into several clips, which creates visual gaps between consecutive video segments, and does not preserve the video context. Most recently, the algorithms on fast-forwarding videos have emerged as effective approaches to deal with the tasks of retrieving meaningful segments without losing the temporal continuity \cite{Lan2018,Okamoto2013,Ramos2016,Silva2018}. On the flip side, fast forward approaches are limited by the lack of a well-defined semantic definition.


This paper takes a step forward towards fast-forwarding videos based on the semantics of the content. By using textual data to guide an agent that seeks the best set of frames to be removed, our method emphasizes highly semantic content segments, while preserving the temporal continuity. We evaluate our approach on the challenging YouCook2 dataset~\cite{Zhou2018Towards}. The experiment has shown that our method achieves the best performance in terms of F1 Score and coverage at the video segment level. 

\paragraph{Contributions.} The contributions of this work can be summarized as follows: \textit{i)} a new fast-forward method based on a reinforcement learning formulation, which is able to accelerate videos according to frame similarity scores with textual data; \textit{ii)} a novel Visually-guided Document Attention Network (VDAN) capable of generating a highly discriminative embedding space for textual and visual data.


\section{Related Work}
\label{sec:related_work}

Various methods have been proposed in the literature to deal with the task of shortening a video using different approaches such as summarization~\cite{Lee2012, Fei2017, Mahasseni2017, Zhang2016}, fast forward~\cite{Poleg2015,Joshi2015,Ramos2016,Silva2018cvpr,Lan2018}, cross-modal techniques~\cite{Plummer2017}, and  reinforcement learning~\cite{Lan2018}. In the following, we present the works most related to ours, as well as the most representative techniques of each approach.

\paragraph*{Video summarization.}

Over the past several years, video summarization methods were the big players when dealing with the task of shortening a video~\cite{Lee2012, Fei2017, Mahasseni2017, Zhang2016}. The shorter version is, usually, a summary of the input video composed of a storyboard of keyframes or video skims with the most distinguishable segments~\cite{Molino2017}. Most of the summarization methods select the frames or skims using a relaxed or non-existent temporal restriction, resulting in visual gaps and breaking the video context.

The strategy adopted by researchers to create the summary varies from clustering visual features of frames~\cite{Mahasseni2017} and training neural networks that infer the representativeness of a video segment~\cite{Yao2016, Zhang2016}, to employing additional information such as user queries, external sensors~\cite{Lee2012}, or textual annotations~\cite{Plummer2017}. Lee~\etal~\cite{Lee2012}, in the context of first-person videos, analyzed properties such as social interaction, gaze, and object detection to create a storyboard summary of the input video. Zhang~\etal~\cite{Zhang2016} proposed a method to create either a storyboard or skims by modeling long-range dependencies among the video frames using a bi-directional Long Short-Term Memory (LSTM) recurrent network. Yao~\etal~\cite{Yao2016} performed the selection of the relevant segments fusing information from spatial and temporal Deep Convolutional Neural Networks (DCNNs) to identify highlighting moments in sports videos.

Reinforcement learning has also been applied to video summarization~\cite{Zhou2018Deep, Lan2018}, motivated by the fact that it had been successfully applied to many challenging tasks, such as mastering complex games like Go~\cite{Silver2017}, Shogi~\cite{Silver2018}, and achieving super-human performance in Atari games~\cite{Hasselt2016}. Additionally, it has great application in vision tasks, including visual tracking~\cite{Yun2017}, and active object recognition~\cite{Paletta2000}. Zhou~\etal~\cite{Zhou2018Deep} presented an end-to-end unsupervised framework also based on the reinforcement learning paradigm. Their method summarizes videos by applying a  diversity-representativeness reward that guides the agent to create more diverse and more representative summaries.  

\paragraph*{Semantic fast-forward.}

The lack of context that emerges from the gaps generated by video summarization methods creates a nuisance to the consumers of instructional videos. The existence of a gap also might confuse the user about the whole process. In other words, the user would be unaware if an important step was missed with the gap if the original video is unknown. Fast-forward based methods add time constraint in the frame sampling, which results in a shorter and contiguous version of the input video.

Some approaches also deal with visual stability constraints when sampling the frames, achieving smooth final videos by modeling the sampling step as an optimization problem~\cite{Poleg2015,Joshi2015,Kopf2014,Ramos2016}. A drawback of applying previous fast-forward methods in instructional videos is that the whole video would be sped-up. A recipe video, for example, might have a long and straightforward step, like boiling vegetables. This step would be sped-up with the same rate as a shorter and complicated task like filleting a fish.

Semantic fast-forward methods, on the other hand, emphasize segments with a high semantic load. The emphasis effect is achieved by accelerating the relevant segments with a lower speed-up rate when compared with the rate applied to the rest of the video. Okamoto and Yanai~\cite{Okamoto2013} proposed a method to fast-forward guidance videos emphasizing video segments containing pedestrian crosswalks or turning movements in street corners. Ramos~\etal~\cite{Ramos2016} presented a semantic fast-forward method for first-person videos dealing with visual stability constraints with emphasis on faces/pedestrians. Silva~\etal~\cite{Silva2018} extended the work of Ramos~\etal, including an automatic parameter setting based on Particle Swarm Optimization (PSO) algorithm, a Convolutional Neural Network to assign frame scores based on Internet users' preferences, and a video stabilizer proper to fast-forwarded videos~\cite{Silva2016}. Silva~\etal~\cite{Silva2018cvpr} proposed a semantic fast-forward to first-person videos by modeling the frame sampling as a Minimum Sparse Reconstruction problem. A drawback of this work is that it requires a pre-processing step that is time-consuming and relies on the accuracy of other methods.

In a recent work, Lan~\etal~\cite{Lan2018} introduced the Fast Forward Net (FFNet). Their methodology summarizes videos on the fly using a reinforcement learning based method to select frames with the most memorable views according to human-labeled data. Similar to FFNet and Zhou~\etal, we also apply an agent that is trained by the reinforcement learning paradigm; however, our approach is a step towards training agents to work in a cross-modal embedding space.

\paragraph*{Cross-modal embedding.}

Recently, the algorithms on cross-modal embedding have emerged as promising and effective approaches to deal with a variety of tasks such as video description~\cite{Pan2016} and text-based image or video retrieval \cite{Mithun2019, Dong2018, Aytar2017, Pan2016}, to name a few. Virtually all these methods rely on creating a shared embedding space, where features from multiple modalities can be compared. 

A successful application of a cross-modal approach has been presented by Plummer~\etal~\cite{Plummer2017}. The authors created a video summary approach that selects the best subset of video segments by analyzing their visual features (\eg, representativeness, uniformity, and interestingness) along with vision-language modeling. Salvador~\etal~\cite{salvador2017im2recipe} applied a multi-modal neural model to learn a common embedding space for images and recipes and tackled the task of retrieving recipes from image queries. Carvalho~\etal~\cite{Carvalho2018} extended the method of Salvador~\etal to use a different loss function. Wang~\etal~\cite{Wang2019} proposed an adversarial learning strategy to align both modalities.

Most of these works perform document retrieval using as query an image representing the final result of a recipe. Our proposed cross-modal embedding (VDAN), on the other hand, provides the semantic distance between each frame in the instructional video and the textual steps described in the document, \ie, the recipe. 


\section{Methodology}
\label{sec:methodology}

\begin{figure*}[t!]
	\centering
	\includegraphics[width=.99\linewidth]{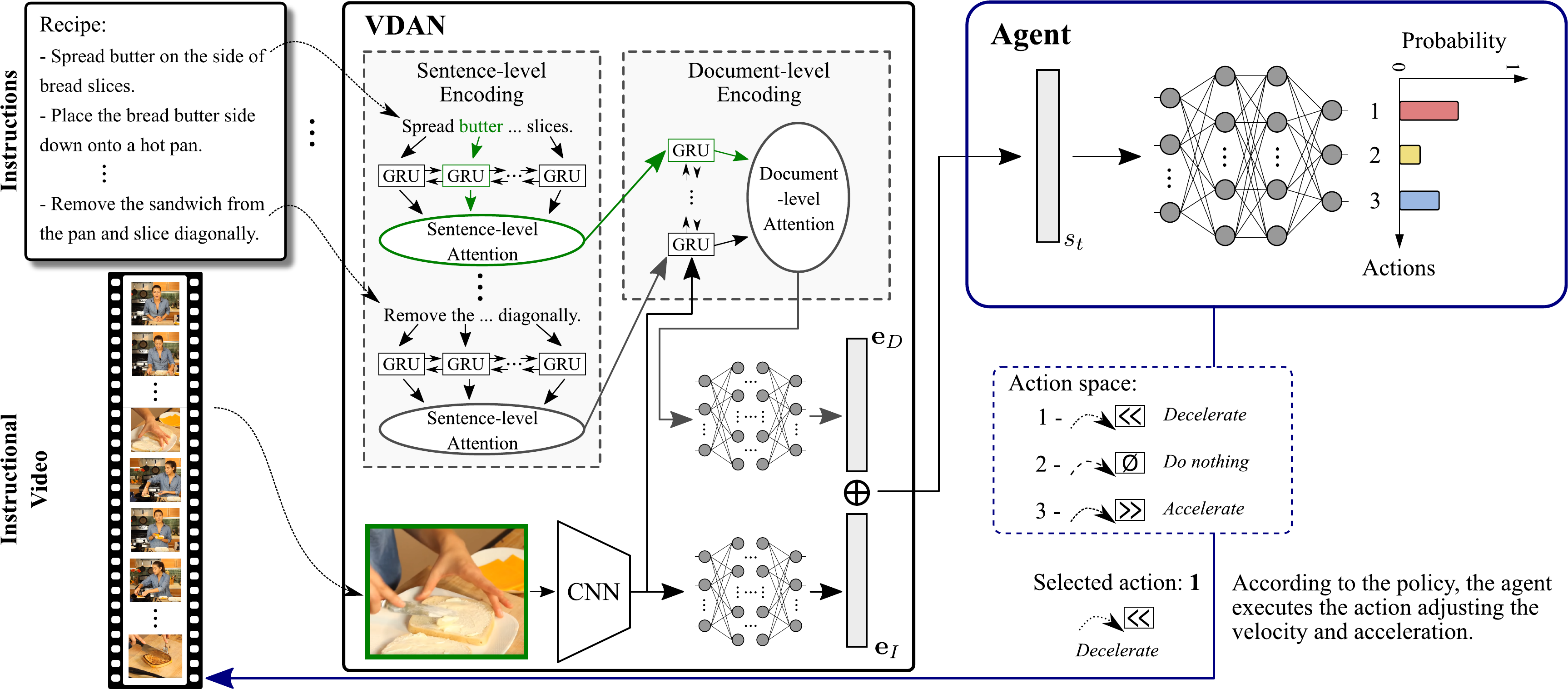}
	\caption{Our methodology is composed of two main stages. First, we employ our Visually-guided Document Attention Network (VDAN) to create a cross-modal embedding that encodes the user document and the input video frames. Then, following the reinforcement learning paradigm, we train an agent to select which frames to remove executing the actions to increase, decrease, or keep the speed-up rate.}
	\label{fig:methodology}
\end{figure*}

Our method is based on an encoding-decoding framework to create fast-forwarded videos. The first stage of our method consists of the novel Visually-guided Document Attention Network (VDAN). The VDAN creates an embedding space for encoding documents and images. In the second stage, we formulate the fast-forwarding task as a sequential decision-making process. We train a reinforcement learning agent that observes the encoded text and video frame and decodes them into a distribution over the actions for increasing, decreasing, or maintaining the current speed-up rate of the output video. Figure~\ref{fig:methodology} illustrates the main steps of our approach.

\subsection{Visually-guided Document Attention Network}

Since our ultimate goal is to create a fast-forward video by dropping non-relevant frames given an input document, we propose the Visually-guided Document Attention Network (VDAN). Our network takes a document and an image as input and, guided by the visual features, creates representative feature vectors for both modalities. By training VDAN, we aim at creating an embedding space in which textual and visual features are aligned. We argue the aligned embedding vectors help our agent make sense of the semantic proximity between a frame and the document, and then learn the best policy to discard non-relevant frames as far as the document is concerned (\eg, a recipe).

Formally, let ${D = \{p_1, p_2, \cdots, p_N\}}$ be the document composed of $N$ sentences, and $I$ be the image that feeds the network. In our task, $D$ is represented by a document composed of a set of textual instructions, and $I$ is a video frame. VDAN produces $d$-dimensional embeddings $\mathbf{e}_D \in \mathbb{R}^d$ and $\mathbf{e}_I \in \mathbb{R}^d$ for textual and visual data, respectively, given the parameters ${\theta_E = \{\theta_D, \theta_I\}}$. 

\paragraph{Document Encoder.} To encode $D$, we employ a Hierarchical Recurrent Neural Network (H-RNN) coupled with a soft attention mechanism in each level~\cite{Yang2016, Zhao2018}, since an H-RNN can capture long-range temporal dependencies~\cite{Zhao2017}. 

Our H-RNN is composed of two levels of encoding: i) the sentence-level and ii) the document-level, as illustrated in Figure~\ref{fig:methodology}. Each level contains bi-directional GRU~\cite{cho2014learning} units that produce hidden state vectors. These vectors feed the attention layer. Let $\mathbf{w}_{i1}, \mathbf{w}_{i2}, \cdots, \mathbf{w}_{iM_{i}}$ denote the distributional word representation~\cite{pennington2014glove} of each word in sentence $p_{i}$. The sentence-level encoder produces a hidden state vector ${\mathbf{h}_{ij} = f_p(\mathbf{w}_{ij}; \mathbf{h}_{i(j-1)}, \theta_{R_p})}$ at each timestep $j$ given the word embedding $\mathbf{w}_{ij}$, the previous hidden state $\mathbf{h}_{i(j-1)}$, and the parameters $\theta_{R_p}$. As stated by Yang~\cite{Yang2016}, words present different contributions to the meaning of a sentence. Therefore, we feed $\mathbf{h}_{ij}$ to the attention module defined as:
\begin{align}
\mathbf{u}_{ij} &= \text{tanh}(W_p\mathbf{h}_{ij}),\\
\alpha_{ij} &= \frac{\text{exp}(\mathbf{u}_{ij}^\intercal c_p)}{\sum_j \text{exp}(\mathbf{u}_{ij}^\intercal c_p)},\\
\mathbf{p}_i &= \sum_j \alpha_{ij}\mathbf{h}_{ij},
\end{align}
\noindent where $\mathbf{u}_{ij}$ is a hidden representation of $\mathbf{h}_{ij}$, $\alpha_{ij}$ gives the importance weights for each $\mathbf{h}_{ij}$, $\mathbf{p}_i$ is the sentence-level embedding for the sentence $p_{i}$, $c_p$ is a word-level context vector which acts as a fixed query to find the informative word, and $W_p$ is a projection matrix. The alignment between $c_p$ and $\mathbf{u}_{ij}$ defines the score used to compute $\alpha_{ij}$.

In the document-level encoding, each $\mathbf{p}_i$ is used to produce a hidden state vector ${\mathbf{h}_{i} = f_d(\mathbf{p}_{i}; \mathbf{h}_{i-1}, \theta_{R_d})}$. Different sentences may also present different contributions to the document. In our approach, the instructional characteristic of the document increases the probability of a given video frame being similar to only one instruction. Thus, similar to the sentence-level counterpart, we also employ an attention module, which is parameterized by $W_d$ and $c_d$. As a result, after feeding the document-level encoder with all vectors $\mathbf{p}_{i}$, it yields the document-level encoding $\mathbf{d}$. Finally, we project $\mathbf{d}$ into the embedding space using a fully connected network $f_D$ parameterized by $\theta_D$. Thus, ${\mathbf{e}_D = f_D(\mathbf{d}; \theta_D)}$.

\paragraph{Image Encoder.} To produce the image embedding $\mathbf{e}_I$, we first extract the image features with a ResNet-50~\cite{he2016deep} encoder, producing an intermediate vector ${\phi(I) \in \mathbb{R}^z}$. Then, we project $\phi(I)$ into the embedding space using a fully connected network $f_I$ parameterized by $\theta_I$ as follows ${\mathbf{e}_I = f_I (\phi(I); \theta_I )}$. To guide the document-level attention module to attend the correct sentence, we set the first hidden state vector of the document-level encoder as ${\mathbf{h}_0 = \phi(I)}$. For the sentence-level, however, we set $ \mathbf{h}_{i0} = \mathbf{0}$.

Both document and image encoders also include an $\ell_2$ normalization layer to make $\mathbf{e}_D$ and $\mathbf{e}_I$ unit norm vectors. The attention module learns to attend the correct words and sentences to produce an embedding ${\mathbf{e}_D}$ more aligned to ${\mathbf{e}_I}$.

\paragraph{Training.} For each image $I$ in the training set, we create a positive and a negative document, $D^+$ and $D^-$, to compose the training pairs $<$$D^+, I$$>$ and $<$$D^-, I$$>$. A positive document $D^+$ consists of the sentences that describe the image $I$ and, additionally, sentences that describe a randomly selected image $I^\prime$. The strategy of adding sentences that do not describe the image helps the document-level attention module to attend the proper sentences at training time. To create the negative document, $D^-$, we randomly select two other images $I^\prime$ and $I^{\prime\prime}$, and collect their respective sentences. At each training step, we shuffle all the sentences in the document for generalization purposes.

In order to create more aligned embeddings, we optimize ${\theta_{enc} = \{\theta_{R_{p}}, W_p, c_p, \theta_{R_{d}}, W_d, c_d, \theta_E, \theta_D\}}$ by minimizing the cosine embedding loss as follows:
\begin{equation}
\mathcal{L}_{enc}(\hat{D}, \hat{I}; \theta_{enc}) = 
\begin{cases}
1 - \cos(\mathbf{e}_D, \mathbf{e}_I), & \text{if}\ y = 1\\
\max(0, \cos(\mathbf{e}_D, \mathbf{e}_I) - \eta), & \text{otherwise},
\end{cases}
\end{equation}
\noindent where $\hat{D}$ and $\hat{I}$ are the training document and image, respectively, $y$ is equal to $1$ if $\hat{D}$ and $\hat{I}$ correspond, and $\eta$ is a margin parameter, which is set to $0$ in our problem.

\subsection{Semantic Fast-Forward Network (SFF-RL)}

In the second stage of our methodology, we define our decoding phase, in which the agent observes the encoded vectors $\mathbf{e}_D$ and $\mathbf{e}_I$ and sample an action over the action space to adjust the speed-up rate accordingly.

We formulate the problem of selecting frames as a Markov Decision Process (MDP). In our formulation, we train an agent to maximize the expected sum of discounted rewards:
\begin{equation}
R_t = \mathbb{E}\left[\sum_{n=0}^{T - t}\gamma^n r_{t+n}\right],
\end{equation}
\noindent where $t$ is the current timestep, $r_{t+n}$ is the reward $n$ timesteps into the future, and $T$ is the total number of timesteps. At each timestep, one frame is selected; therefore, $t$ also indicates the current number of the selected frames. ${\gamma \in (0,1]}$ is a discount factor. In our case, however, future rewards are equally important, and hence we use ${\gamma = 1}$.

In this problem, the agent observes a video and a text, and must take actions to create an optimal accelerated version of the input video. Since we want to keep the overall coherence of the video instead of trimming it, we model an agent that navigates in the video space. \Ie, the agent has velocity $v$ and acceleration $\omega$ and based on the current velocity, the next frame is selected. Therefore, the agent goes through the whole video, but skips frames according to a dynamically changing velocity. At each timestep, the agent can increase, decrease, or keep its current acceleration, which will, in turn, affect the velocity. Since we apply Model-free Reinforcement Learning, the transition function does not need to be pre-defined, nor learned; as the agent focus directly on learning the best policy. In the following, we define all elements used in our MDP formulation.

\paragraph{State and actions.} In order to allow an agent to effectively navigate through the video space, we define the state vector as the concatenation of the document and frame embeddings, \ie, ${\mathbf{s}_t = [\mathbf{e}_D; \mathbf{e}_I] \in \mathcal{S}}$. To create a fast-forwarded video using a textual input, our agent adaptively adjusts the speed-up rate such that the video segments which semantically match the input text are exhibited in a lower speed and the others in a higher speed. Thus, the agent's action space $\mathcal{A}$ has three actions: i) \emph{decelerate}; ii) \emph{do nothing}; and iii) \emph{accelerate}. As mentioned, the agent has a current velocity $v$, and hence would skip the next $v$ frames for whichever action it takes. \emph{Decelerate} and \emph{accelerate} update the velocity and acceleration states of the agent as ${v = v - \omega}$ and ${\omega = \omega - 1}$ for \emph{decelerate}, and ${v = v + \omega}$ and ${\omega = \omega + 1}$ for \emph{accelerate}, while \emph{do nothing} keeps the current $v$ and $\omega$. Additionally, acceleration and velocity saturate at certain values $\omega_{max}$ and $v_{max}$, respectively, and they are always greater than or equal to $1$. Note that $ \omega $ does not correspond to a physical acceleration, allowing the agent to quickly adjust the velocity to collect more rewards when the semantic level changes. 



\paragraph{Reward function.} The goal of the agent is to learn a policy ${\pi(a | s_t, \theta_\pi)}$, which represents the probability of the agent taking a certain action ${a \in \mathcal{A}}$ given the state $s_t$ and the parameters $\theta_\pi$. The reward should encourage the agent to increase or decrease the speed-up of the video, given the semantic similarity between the visual and textual data in the upcoming frame. Therefore, we design an immediate reward proportional to the text and frame features alignment. Thus, at training time, after taking the action ${a_t \sim \pi(a | s_t, \theta_\pi)}$ in the $t^{th}$ step, the agent receives the following reward signal:
${ r_t = \mathbf{e}_D \cdot \mathbf{e}_I}$.

Note that the agent receives higher rewards if $\mathbf{e}_D$ and $\mathbf{e}_I$ point to the same direction in the embedding space, which encourages it to reduce the speed and accumulate more rewards since the temporal neighboring frames are more likely to yield higher reward values.

\paragraph{Objective.} Apart from aligning the textual and visual features produced by VDAN, the overall objective of our framework also tries to maximize the expected cumulative reward $R_t$ at each timestep $t$. We follow the REINFORCE algorithm~\cite{Williams1992} to learn the parameters $\theta_\pi$ in order to maximize the expected utility 
\begin{equation}
J(\theta_\pi) = \sum_{a \in \mathcal{A}} \pi(a | s_t, \theta_\pi) R_t.
\end{equation} 
In order to improve the learning performance, we employ the advantage function approach~\cite{Sutton2018}, and maximize the expected advantage:
\begin{equation}
J^\prime (\theta_\pi) = \sum_{a \in \mathcal{A}} \pi(a | s_t, \theta_\pi)(R_t - v(s_t | \theta_v)),     
\end{equation}
\noindent where $v (s_t | \theta_v)$ is a function parameterized by $\theta_v$, which predicts our expected cumulative reward at state $s_t$. The gradient of $J^\prime$, $\nabla_{\theta_\pi}J^\prime(\theta_\pi)$, is given by
\begin{equation}
\sum_{a \in \mathcal{A}} \pi(a | s_t, \theta_\pi) (\nabla_{\theta_\pi} \log \pi(a | s_t, \theta_\pi)) (R_t - v(s_t|\theta_v)). 
\end{equation}
Usually, Monte Carlo sampling is applied, due to the high dimension of the action sequence space, leading to the following approximation for the gradient: 
\begin{equation}
\nabla J^\prime(\theta_\pi) \approx \sum_t \nabla_{\theta_\pi} \log \pi (a_t | s_t, \theta_\pi) (R_t - v(s_t|\theta_v)),
\end{equation}
\noindent where $a_t$ is the action taken at time $t$. Hence, we minimize the following loss function:
\begin{equation}
\mathcal{L}^\prime (\theta_\pi) = - \sum_t \left(\log \pi(a_t | s_t, \theta_\pi)\right) (R_t - v(s_t | \theta_v)).
\end{equation}
Also, it is usually recommended to add the entropy of the policy output $H(\pi(a_t | s_t,\theta_\pi))$ into the loss, in order to have a greater action diversity~\cite{Li2018}. Therefore, our final policy loss is given by
\begin{equation}
\mathcal{L}_{dec} (\theta_\pi) = \mathcal{L}^\prime(\theta_\pi) - \sum_t \beta \cdot H(\pi(a_t|s_t,\theta_\pi)),
\end{equation}
\noindent where $\beta$ is a constant to balance the entropy importance. In our experiments, we set $\beta$ to $0.01$.

Additionally, we also need to learn the state-value function $v(s_t | \theta_v)$. We do that by minimizing the mean squared error:
\begin{equation}
\mathcal{L}_v (\theta_v) = \sum_t \left(v(s_t | \theta_v) - R_t \right)^2.
\end{equation}
Both losses $\mathcal{L}_v$ and $\mathcal{L}_{dec}$ can now be minimized using stochastic gradient descent.

At test time, we use $ \argmax_a \pi(a|s_t,\theta_\pi)$ as the chosen action for the agent in a given timestep $t$.


\section{Experiments}
\label{sec:experiments}

In this section, we investigate the performance of our method evaluating it both qualitatively and quantitatively on different recipe videos.


\subsection{Experimental Setup}
\paragraph{Dataset and Evaluation Metric.} We extract a subset of videos from the YouCook2 dataset~\cite{Zhou2018Towards} to compose the sets for training and testing our approach. The videos in the dataset were collected from YouTube and are distributed over $89$ recipes such as {\it grilled cheese}, {\it hummus}, \etc. Each video has up to $16$ English sentences localized by timestamps, and each of these sentences corresponds to a video segment in which the instruction is being executed. Since we aim at creating shorter videos that convey the same information of the original video, we only evaluated our method in videos whose instruction segments correspond to at most $25\%$ of its length, resulting in a total of $121$ videos.

To evaluate the performance of each method, we computed the F1 Score and, following Gygli~\etal~\cite{gygli2015cvpr} and Lan~\etal~\cite{Lan2018}, we also evaluate our method in terms of coverage at video segment level. While the F1 Score consists of a weighted average of Precision and Recall, the coverage at video segment level gives the quality of the coverage of frames manually annotated as relevant. We consider that an important segment has been taken if the number of relevant frames (according to the ground truth) selected by an approach is higher than a threshold, the hit number.

\begin{table*}[!t]
	\centering
	\setlength{\tabcolsep}{7pt}
	\begin{tabular}{@{}lrrrcrrrcrrr@{}}
		\toprule
		\multicolumn{1}{c}{\multirow{2}{*}{Test Set}} & \multicolumn{3}{c}{\textbf{Precision}} & & \multicolumn{3}{c}{\textbf{Recall}} & & \multicolumn{3}{c}{\textbf{F1 Score}} \\  
		& \multicolumn{1}{c}{SSFF} & \multicolumn{1}{c}{FFNet} & \multicolumn{1}{c}{Ours} & & \multicolumn{1}{c}{SSFF} & \multicolumn{1}{c}{FFNet} & \multicolumn{1}{c}{Ours} & & \multicolumn{1}{c}{SSFF} & \multicolumn{1}{c}{FFNet} & \multicolumn{1}{c}{Ours} \\ 
		\cmidrule(){2-4} \cmidrule(){6-8} \cmidrule(){10-12}
		Waldorf salad       & $          0.20 $ & $          0.19 $ & $ \mathbf{0.34} $ &  & $          0.15 $ & $ 0.04 $ & $ \mathbf{0.72} $ &  & $          0.17 $ & $ 0.07 $ & $ \mathbf{0.46} $ \\
		Grilled cheese      & $          0.22 $ & $          0.23 $ & $ \mathbf{0.27} $ &  & $          0.16 $ & $ 0.04 $ & $ \mathbf{0.39} $ &  & $          0.19 $ & $ 0.07 $ & $ \mathbf{0.32} $ \\
		Corn dogs           & $          0.14 $ & $          0.17 $ & $ \mathbf{0.18} $ &  & $          0.08 $ & $ 0.05 $ & $ \mathbf{0.20} $ &  & $          0.10 $ & $ 0.08 $ & $ \mathbf{0.19} $ \\
		Hash browns         & $          0.27 $ & $          0.22 $ & $ \mathbf{0.32} $ &  & $          0.16 $ & $ 0.04 $ & $ \mathbf{0.29} $ &  & $          0.20 $ & $ 0.07 $ & $ \mathbf{0.31} $ \\
		Bangers and mash    & $          0.21 $ & $          0.29 $ & $ \mathbf{0.32} $ &  & $          0.11 $ & $ 0.12 $ & $ \mathbf{0.69} $ &  & $          0.15 $ & $ 0.17 $ & $ \mathbf{0.44} $ \\
		Foie gras*          & $          0.18 $ & $ \mathbf{0.21} $ & $          0.17 $ &  & $          0.16 $ & $ 0.13 $ & $ \mathbf{0.42} $ &  & $          0.16 $ & $ 0.15 $ & $ \mathbf{0.24} $ \\
		Escargot            & $ \mathbf{0.33} $ & $          0.23 $ & $          0.30 $ &  & $ \mathbf{0.26} $ & $ 0.08 $ & $ \mathbf{0.26} $ &  & $ \mathbf{0.29} $ & $ 0.12 $ & $          0.28 $ \\
		Sauerkraut*         & $ \mathbf{0.28} $ & $          0.22 $ & $          0.25 $ &  & $          0.19 $ & $ 0.09 $ & $ \mathbf{0.37} $ &  & $          0.23 $ & $ 0.13 $ & $ \mathbf{0.30} $ \\
		Goulash             & $          0.20 $ & $ \mathbf{0.30} $ & $          0.27 $ &  & $          0.14 $ & $ 0.13 $ & $ \mathbf{0.87} $ &  & $          0.17 $ & $ 0.18 $ & $ \mathbf{0.41} $ \\
		Beef bourguignon    & $          0.25 $ & $          0.25 $ & $ \mathbf{0.26} $ &  & $          0.11 $ & $ 0.05 $ & $ \mathbf{0.30} $ &  & $          0.16 $ & $ 0.09 $ & $ \mathbf{0.28} $ \\
		Wiener schnitzel    & $          0.28 $ & $          0.23 $ & $ \mathbf{0.29} $ &  & $          0.17 $ & $ 0.06 $ & $ \mathbf{0.24} $ &  & $          0.21 $ & $ 0.09 $ & $ \mathbf{0.26} $ \\
		Pasta e fagioli*    & $          0.24 $ & $ \mathbf{0.43} $ & $          0.33 $ &  & $          0.12 $ & $ 0.22 $ & $ \mathbf{0.54} $ &  & $          0.16 $ & $ 0.29 $ & $ \mathbf{0.41} $ \\
		Hummus              & $          0.30 $ & $          0.25 $ & $ \mathbf{0.52} $ &  & $          0.19 $ & $ 0.05 $ & $ \mathbf{0.95} $ &  & $          0.23 $ & $ 0.09 $ & $ \mathbf{0.67} $ \\
		Udon noodle soup    & $ \mathbf{0.22} $ & $          0.18 $ & $          0.11 $ &  & $ \mathbf{0.18} $ & $ 0.04 $ & $          0.05 $ &  & $ \mathbf{0.20} $ & $ 0.07 $ & $          0.07 $ \\
		Indian lamb curry*  & $          0.17 $ & $ \mathbf{0.23} $ & $          0.16 $ &  & $          0.11 $ & $ 0.13 $ & $ \mathbf{0.16} $ &  & $          0.13 $ & $ 0.16 $ & $ \mathbf{0.16} $ \\
		Dal makhani         & $          0.22 $ & $ \mathbf{0.38} $ & $          0.20 $ &  & $          0.14 $ & $ 0.13 $ & $ \mathbf{0.23} $ &  & $          0.17 $ & $ 0.19 $ & $ \mathbf{0.22} $ \\
		Wanton noodle       & $ \mathbf{0.21} $ & $          0.19 $ & $          0.20 $ &  & $          0.15 $ & $ 0.09 $ & $ \mathbf{0.96} $ &  & $          0.17 $ & $ 0.12 $ & $ \mathbf{0.33} $ \\
		Masala dosa         & $          0.11 $ & $          0.14 $ & $ \mathbf{0.16} $ &  & $          0.08 $ & $ 0.10 $ & $ \mathbf{0.74} $ &  & $          0.09 $ & $ 0.12 $ & $ \mathbf{0.27} $ \\ 
		\cmidrule(){2-12}
		{\bf Mean}                & $ \mathit{ 0.22 }  $ & $ \mathit{ 0.24 } $ & $ \mathbf{ 0.26 } $ &   & $ \mathit{ 0.15 } $ & $ \mathit{ 0.09 } $ & $ \mathbf{ 0.47 } $ &   & $ \mathit{ 0.18 } $ & $ \mathit{ 0.12 } $ & $ \mathit{ \mathbf{0.31}  } $ \\
		{\bf Std}  & $ \mathit{ 0.06 } $ & $ \mathit{ 0.07 } $ & $ \mathit{ 0.09 } $ &   & $ \mathit{ 0.04 } $ & $ \mathit{ 0.05 } $ & $ \mathit{ 0.29 } $ &   & $ \mathit{ 0.05 } $ & $ \mathit{ 0.06 } $ & $ \mathit{ 0.13 } $ \\
		\bottomrule
	\end{tabular}
	\caption{Precision, Recall and F1 Score results for our test set. The * symbol indicates recipes for which we have collected two recipes and reported their average values. Our method outperforms the baseline competitors in most cases. The best results are in bold.}
	\label{tab:f1_scores}
\end{table*}

\paragraph{Baselines.} We pit our method against the FFNet~\cite{Lan2018} and the work of Silva~\etal~\cite{Silva2018cvpr} on sparse adaptive sampling (SSFF). While SSFF holds state of the art in semantic fast-forward, FFNet, similar to our approach, is a fast-forward method based on the reinforcement learning paradigm.

\paragraph{Implementation Details.}

In our experiments, we use the MSCOCO dataset~\cite{Lin2014} to compose positive and negative pairs to train the VDAN. MSCOCO contains $113{,}287$ training images with $5$ captions each, and $5{,}000$ images respectively for validation and testing. For the VDAN, we use the glove embeddings set pre-trained in the Wikipedia 2014 and Gigaword 5 sets provided by Pennington~\etal~\cite{pennington2014glove}. We set ${d = 128}$ as the dimension of the embedding space and the size of the hidden state vectors $\mathbf{h}_{ij}$ and $\mathbf{h}_i$ to be $1024$ and $2048$, respectively. $f_I$ and $f_D$ are implemented as two independent fully connected neural networks composed of a single hidden layer with $512$ neurons. We train VDAN for $30$ epochs with a batch size of $64$ and obtain the model that had the best performance on validation. The policy network $\pi(a_t|s_t,\theta_\pi)$ and the value-state function $v (s_t | \theta_v)$ are implemented as two independent neural networks with two hidden layers composed of $256$ and $128$ neurons, respectively. We trained our agent during $100$ epochs. Both the VDAN network and the SFF-RL are trained using Adam with a learning rate of ${0.00001}$ for optimization. The value-state approximator, however, is trained with a learning rate of $0.001$ for faster convergence. We set $\omega_{max}$ and $v_{max}$ as $5$ and $20$, respectively. We trained the Lan~\etal's agent with the same number of epochs as ours, with exploration decay ${\epsilon = 0.0001}$. The other parameters are set, as suggested by the authors. Our approach is fully implemented in the PyTorch library, and the experiments were conducted in a single NVIDIA GeForce GTX 1080Ti GPU. 
\begin{figure}[t!]
	\centering
	\includegraphics[width=\columnwidth]{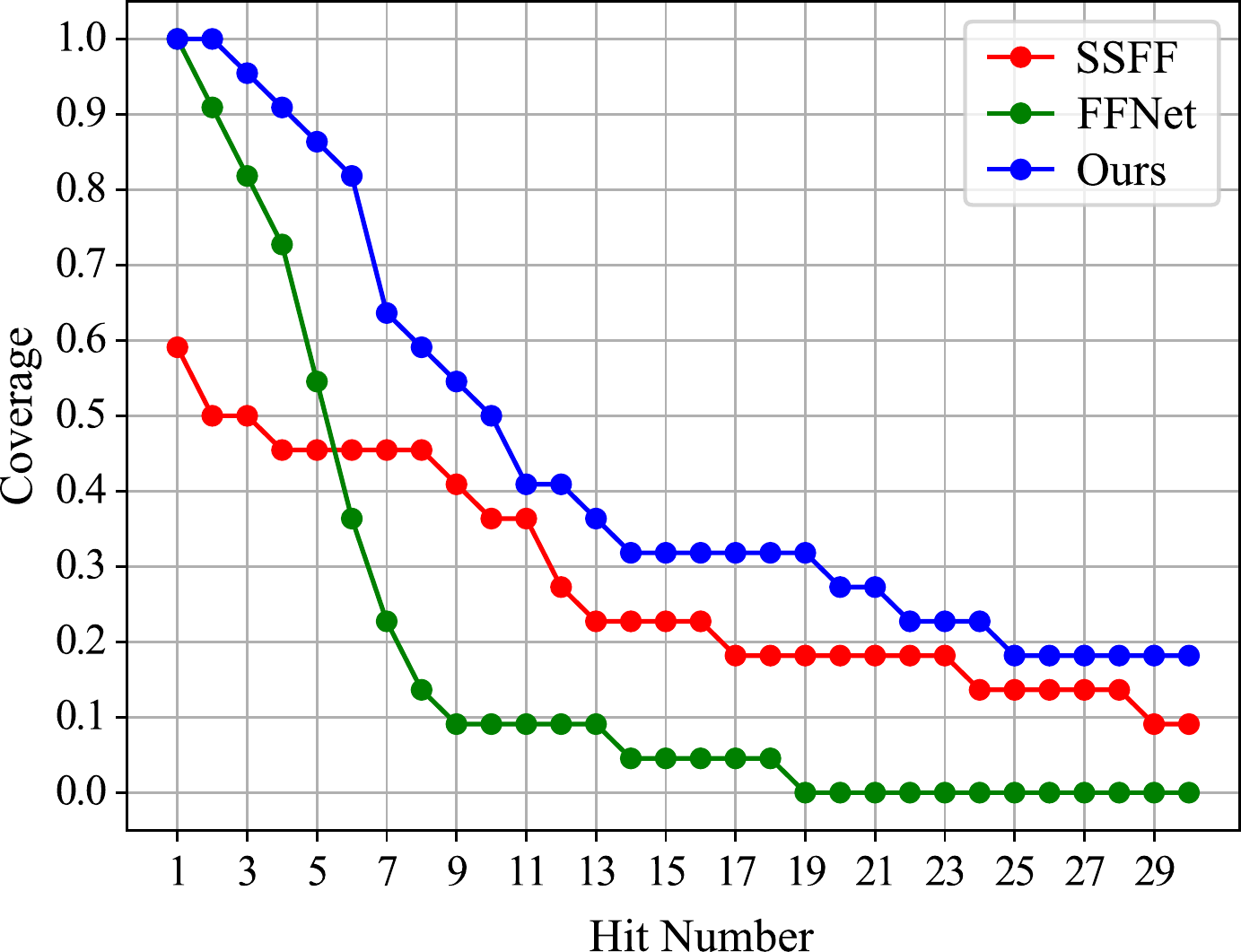}
	\caption{Comparison of the segment-level coverage on our test set. Our method outperforms the competitors in all hit numbers.}
	\label{fig:coverage_results}
\end{figure}
\subsection{Results}

\paragraph{Quantitative results.} Table~\ref{tab:f1_scores} shows the results in terms of precision, recall, and F1 Score in our test set. The recipes marked with a * symbol present two videos; therefore, we computed the mean value of such videos to present in the table. Results for Precision, Recall, and F1 Score show that our method achieves a better performance in comparison to state-of-the-art techniques. We merit the generalization capacity of our agent for the higher recall values presented. At test time, our agent successfully considered the frame and the document to be related. Then, it reduced the speed and acceleration to capture as most relevant frames as it can while maintaining a high precision. Note that our method also outperforms the competitors in terms of precision in several cases. A notable exception is the video of the ``Udon noodle soup'' recipe. In this case, our method achieved an F1 Score of $0.07$. The reason for such a score is that the vectors produced by VDAN along the segments with the instructions were not well aligned in the embedding space, making the agent to \emph{accelerate} or even \emph{decelerate} for a short period of time. This case is illustrated in the Figure~\ref{fig:qualitative_evaluation_failure_case}.

Figure~\ref{fig:coverage_results} depicts the coverage results at segment level. Each point in the plot represents the coverage at a certain hit number. Our method achieves the best performance considering all hit numbers, covering approximately $20\%$ of the important segments when using the higher threshold.
\begin{figure*}[t!]
	\centering
	\includegraphics[width=.98\textwidth]{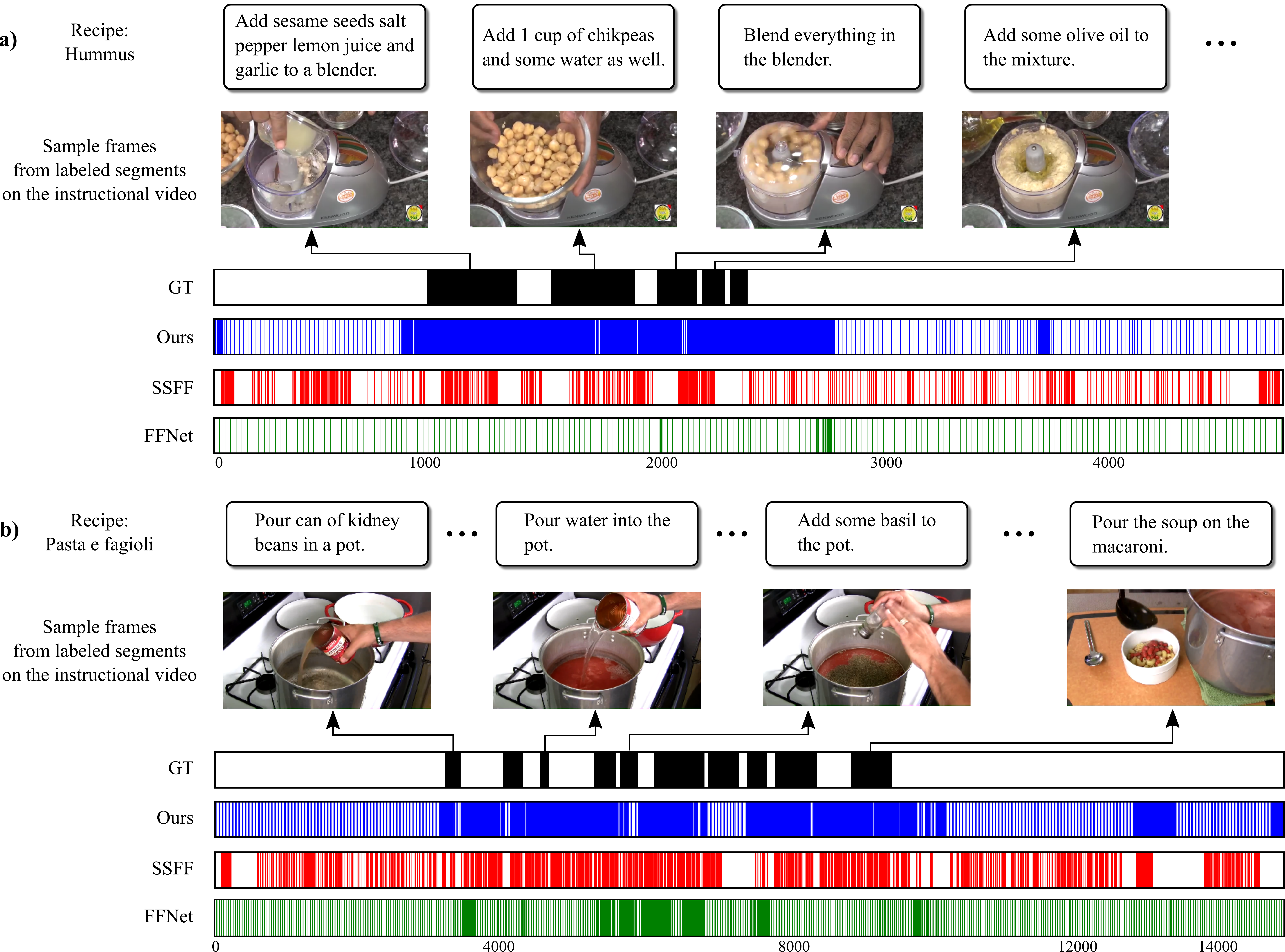}
	\caption{Qualitative results for the compared methodologies in the video for the recipe ``Humus''. The colored bars represent the frame selection produced by the methods, and the black contiguous blocks represent the ground-truth segments. Note that our agent performs a denser sampling in the video segments related to the recipe in both cases a) and b).}
	\label{fig:qualitative_evaluation}
\end{figure*}
\begin{figure}[t!]
	\centering
	\includegraphics[width=\columnwidth]{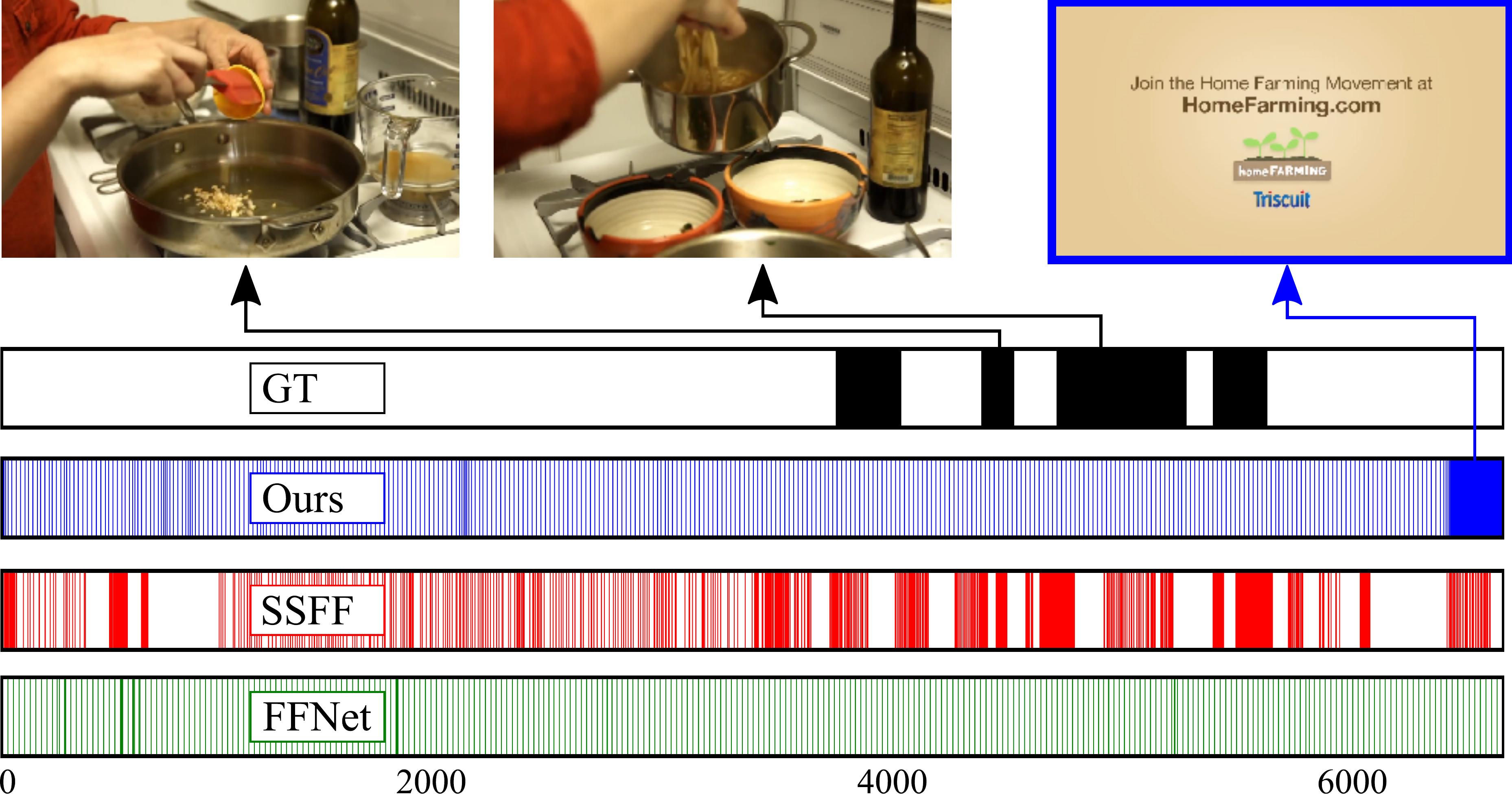}
	\caption{Failure case of our methodology. The colored bars represent the frame selection produced by the methods, and the contiguous black blocks represent the ground-truth segments. VDAN produced misaligned vectors for this video, which guided the agent to a poor selection.}
	\label{fig:qualitative_evaluation_failure_case}
\end{figure}
\paragraph{Qualitative results.} We present our qualitative results in Figure~\ref{fig:qualitative_evaluation}. The colored bars represent the frames selected by each method, while the contiguous black blocks represent the ground-truth segments. The frames and their associated instructions are depicted above the bars. Note that our method presents a denser frame selection when the recipe instructions are shown in the video, in both cases, which indicates that the agent learned a reasonable policy. \Ie, by observing the recipe and the video frame, the agent acts correctly when navigating through the video.

\paragraph{Ablation Study.}
To verify that the word-level attention contributes to the success of the embedding space created by VDAN, we computed the average of the distribution of cosine distances between corresponding and non-corresponding pairs of images and documents. We observed that when adding the word-level attention, the values change from $0.804$ to $0.807$ for the corresponding pairs, and from $0.007$ to $0.006$ for the non-corresponding ones. Due to the slight improvement, we used both the word and sentence-level attention in all experiments.


\section{Conclusions}
\label{sec:conclusions}

In this paper, we proposed a novel methodology based on a reinforcement learning formulation to accelerate instructional videos, where the agent is trained to decide which frames to remove based on textual data. We also presented a novel network called Visually-guided Document Attention Network (VDAN) that creates a highly discriminative embedding space to represent both textual and visual data. Our approach outperforms both FFNet and SSFF methods in terms of F1 Score and coverage at video segment level.


\paragraph{Acknowledgments.} We thank the agencies CAPES, CNPq, FAPEMIG, and Petrobras for funding different parts of this work. We also thank NVIDIA Corporation for the donation of a TITAN Xp GPU.


{\small
\bibliographystyle{ieee_fullname}
\bibliography{2020_cvpr_ramos_arxiv}
}

\clearpage
\setboolean{@twoside}{false}


\section*{Supplementary material}

# [Supplementary Material]
# Straight to the Point: Fast-forwarding Videos via Reinforcement Learning Using Textual Data

Washington Ramos[1]   Michel Silva[1]   Edson Araujo[1]   Leandro Soriano Marcolino[2]
Erickson Nascimento[1]
[1]Universidade Federal de Minas Gerais (UFMG), Brazil    [2]Lancaster University, UK
[1]{washington.ramos, michelms, edsonroteia, erickson}@dcc.ufmg.br, [2]l.marcolino@lancaster.ac.uk

In this supplementary material, we present several qualitative results on recipe videos of the YouCook2 dataset in addition to the ones presented in Figure 4 and Figure 5 of the main paper. In the results, we show the coverage of the selected frames regarding our method and the competitors, *i.e.*, SSFF [2] and FFNet [1]. For each result, we also report the annotated segments.

The results presented in Figure 1 show that the deceleration profile of our method matches the annotated segments of the video. Note that several of the emphasized regions are not perfectly aligned to the ground-truth data. However, it is worth mentioning that there is an emphasized region for each annotated segment of the recipe. The competitors, on the other hand, could not provide the same result.

Even though the FFNet method had been trained in this domain (*i.e.*, using the annotation provided by the dataset), it was capable of only emphasizing a small portion of the annotated segment starting around the frame $3,000$. Regarding the SSFF coverage, it is noteworthy the amount and magnitude of temporal gaps in the frame selection. As pointed before, temporal gaps usually lead to visual discontinuity in the final video. One example is the annotated segment starting around the frame $6,000$, the frame selection of our method emphasized three large segments, while the SSFF skipped almost the entire segment.

Figure 2 depicts the coverage of the selected frames composing the accelerated video generated by our methodology and the two competitors regarding the annotated segments. We can observe that between frames $4,000$ and $4,400$, both SSFF and our method decelerate. In the video used in this experiment, the segment portrays the final dish. Since the visual content of the final dish comprises much of the visual content from the recipe's ingredients, it is expected that our method might interpret such frames as relevant – the SSFF method assigned relevance due to the presence of kitchen-related objects. The SSFF method results show more accentuated gaps between frames, resulting in a visual discontinuity in the final video. It is noteworthy that even though FFNet had been trained using the ground truth annotations from the training dataset, the method failed to emphasize frames that contain the recipe instructions.

In Figure 3, when analyzing the first ground truth segment (GT) of the video, we can visualize that our method was able to emphasize frames in both ends of the segment, but not in the segment itself. By analyzing this annotated portion in the original video (see Figure 3-bottom), we see that the frames where our method decelerates, *i.e.*, those surrounding the ground truth segment, depict the actual step of the recipe. On the other hand, the frames included in the ground truth segment itself do not visually represent the instruction, since they are mainly composed of the executor in close-up. We argue that the densely sampled region of the SSFF method can be explained by the frames inside the GT segment having some visual clues that may lead the YOLO extractor to associate them to the kitchen environment, *e.g.*, microwave, bowl, sink, spoon, *etc*.

The third and fourth ground truth segments of the video are annotated as spaced actions, as showed in Figure 3; however, in the original video, they are presented as consecutive frames in a non-cut video clip. The results show that our method emphasizes the entire segment. The SSFF also performed a dense sampling in these segments; however, it is due to the presence of the oven and the object misclassification of the pan as a bowl. It is worth noting that the SSFF method only analyzes the presence of the relevant objects in the scene. In contrast, our method relates visual information of the frames and text from the input instructions document.

1

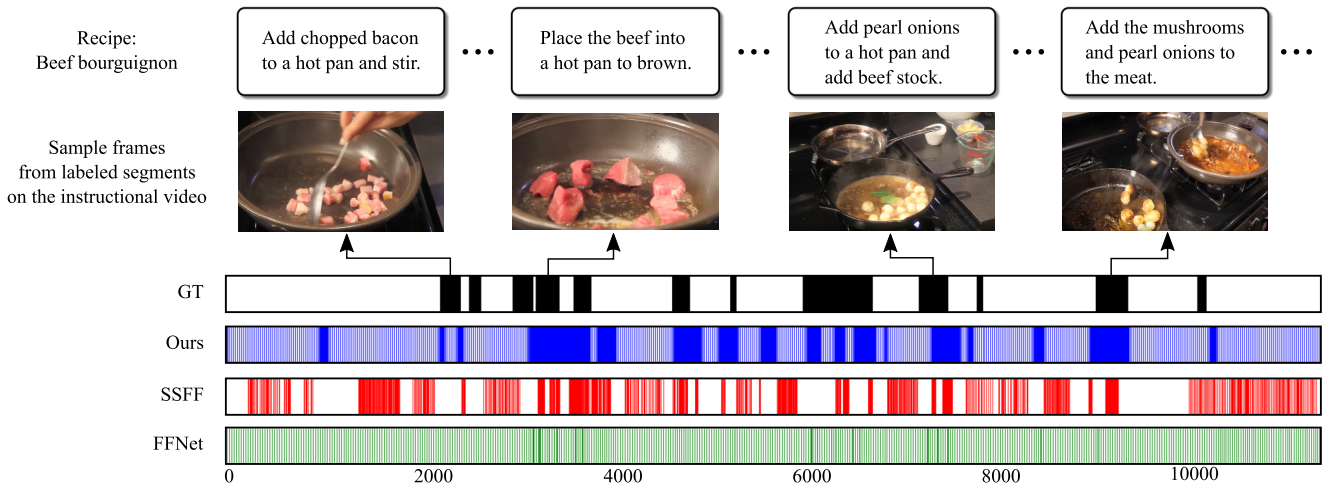

Figure 1. Qualitative results for the compared methodologies in the video for the recipe "Beef Bourguignon" from the YouCook2 dataset. The vertical bars inside the rectangles indicate the selected frames for each method. GT stands to ground truth, and the contiguous black blocks indicate the annotated video segment. The competitors used in our experiments are SSFF and FFNet. In general, our selected frames match most frames from ground truth data.

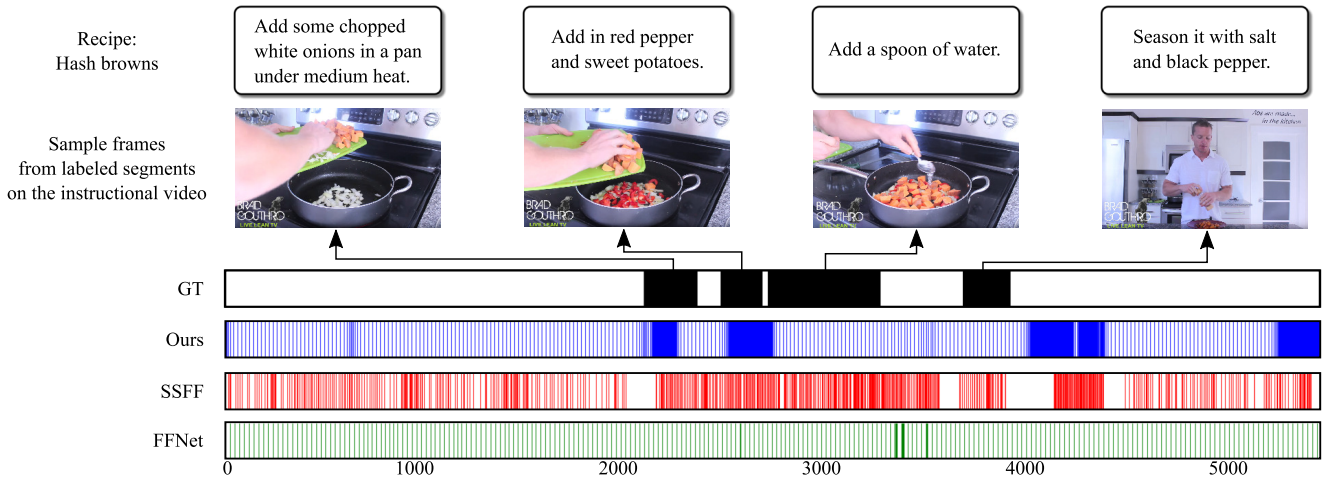

Figure 2. Qualitative results for the compared methodologies in the video for the recipe "Hash Browns" from the YouCook2 dataset. The vertical bars inside the rectangles indicate the selected frames for each method. GT stands to ground truth, and the contiguous black blocks indicate the annotated video segment. The competitors, SSFF and FFNet, present a poor frame selection in terms of GT coverage in comparison to ours.

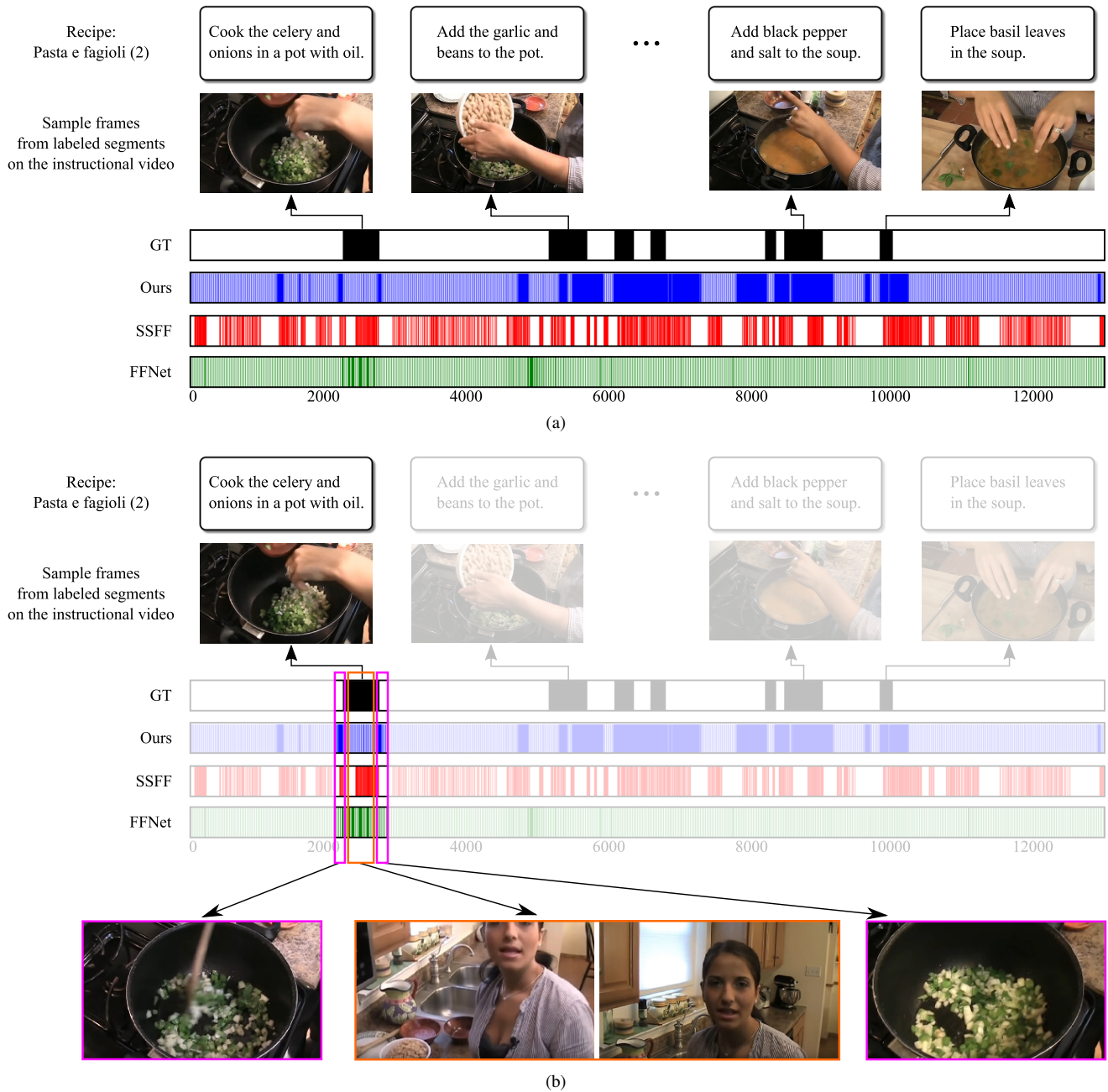

Figure 3. Qualitative results for the compared methodologies in one of the videos for the recipe "Pasta e Fagioli" from the YouCook2 dataset. The vertical bars inside the rectangles indicate the selected frames for each method. GT stands to ground truth, and the contiguous black blocks indicate the annotated video segment. The competitors used in our experiments are SSFF and FFNet. Analyzing the SSFF frame sampling, we notice the temporal gaps resulted from the adaptive frame sampling performed by this technique. The highlighted region in (b) shows that, in the first annotated segment, only the very beginning and the final of this segment shows the recipe. The majority of the frames shows a close-up of the instructor narrating details and curiosities about the original Italian recipe.



\end{document}